# Understanding Reinforcement Learning Algorithms: The Progress from Basic Q-learning to Proximal Policy Optimization

Mohamed-Amine Chadi[a]*, Hajar Mousannif[a]

[a]*Laboratory of informatics systems engineering, department of computer science, faculty of sciences Semlalia, University of Cadi Ayyad, Marrakech 40000, Morocco*

\* Correspondence: mohamedamine.chadi@ced.uca.ma

**Abstract**

This paper presents a review of the field of reinforcement learning (RL), with a focus on providing a comprehensive overview of the key concepts, techniques, and algorithms for beginners. RL has a unique setting, jargon, and mathematics that can be intimidating for those new to the field or artificial intelligence more broadly. While many papers review RL in the context of specific applications, such as games, healthcare, finance, or robotics, these papers can be difficult for beginners to follow due to the inclusion of non-RL-related work and the use of algorithms customized to those specific applications. To address these challenges, this paper provides a clear and concise overview of the fundamental principles of RL and covers the different types of RL algorithms. For each algorithm/method, we outline the main motivation behind its development, its inner workings, and its limitations. The presentation of the paper is aligned with the historical progress of the field, from the early 1980s Q-learning algorithm to the current state-of-the-art algorithms such as TD3, PPO, and offline RL. Overall, this paper aims to serve as a valuable resource for beginners looking to construct a solid understanding of the fundamentals of RL and be aware of the historical progress of the field. It is intended to be a go-to reference for those interested in learning about RL without being distracted by the details of specific applications.

## 1. Introduction

Reinforcement learning (RL) is a branch of machine learning that focuses on training agents to make decisions in an environment to maximize a reward signal [1]. Despite its growing popularity and success in a variety of applications, such as games [2–4], healthcare [5–7], finance [8], and robotics [9–11], RL can be a challenging field for beginners to understand due to its unique setting, jargon, and seemingly more complex mathematics compared to other areas of artificial intelligence. Moreover, many of the existing review papers on RL are application-focused, like the ones presented previously, as well as others [12–14]. This can be distracting for readers who are primarily interested in understanding the algorithms themselves. The only works we know that had focused solely on the RL algorithms are [15–17], however, since they were published in 1996 and 2017 respectively, they did not include many of the classic algorithms that appeared afterward.

To address these challenges, this paper presents a review of the field of RL with a focus on providing a comprehensive overview of the key concepts, techniques, and algorithms for beginners. Unlike other review papers that are application-focused, this paper aims to provide a clear and concise overview of the fundamental principles of RL, without the distractions of specific applications. It also discusses the motivation, inner workings, and limitations of each RL algorithm. In addition, the paper traces the historical progress of the field, from the early 1980s Q-learning algorithm [18] to the current state-of-the-art algorithms, namely, deep Q-learning (DQN) [19], REINFORCE [20], deep deterministic policy gradient (DDPG) [21], twin delayed DDPG (TD3) [22], and proximal policy optimization (PPO) [23].

Overall, this paper aims to serve as a valuable resource for beginners looking to construct a solid understanding of the fundamentals of RL and be aware of the historical progress of the field. By providing a clear and concise introduction to the key concepts and techniques of RL, as well as a survey of the main algorithms that have been developed over the years, this paper aims to help readers overcome the intimidation and difficulties often encountered when learning about RL.

The rest of this paper is organized as follows. In the next section, we provide a brief overview of the general pipeline of an RL setting as well as the key concepts and vocabulary. We then cover the different types of RL algorithms, including value-based, and policy-based methods. In the following section, we trace the historical progress of the field, starting with the early Q-learning algorithm until current state-of-the-art algorithms such as TD3 and PPO.

## 2. Common background

### 1. The general pipeline

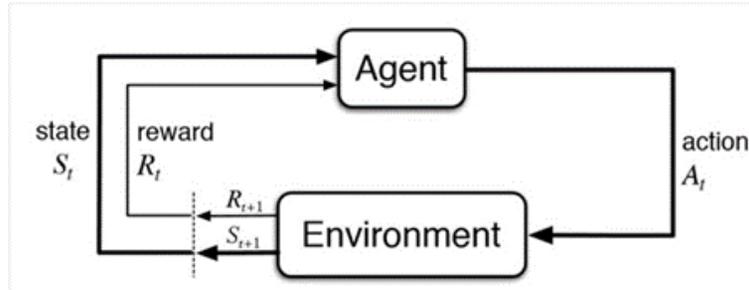

*Figure 1: The general framework in RL*

In the RL pipeline, as illustrated in Figure 1, the agent receives observations from the environment and takes actions based on these observations. The environment responds to the actions by providing a reward signal and transitioning to a new state. The agent uses this feedback to update its policy, which is a mapping from observations to actions. The goal of the agent is to learn a policy that maximizes the long-term return, which is the sum of the rewards received over time.

To learn a good policy, the agent must **explore** the environment and try different actions to learn which actions lead to higher rewards. At the same time, the agent must also **exploit** its current knowledge by taking actions that are expected to lead to the highest reward based on its current policy. This balance between exploration and exploitation is known as **the exploration-exploitation trade-off** and is a key challenge in RL. Overall, the general pipeline of RL involves an agent interacting with an environment, receiving feedback in the form of a reward signal, and updating its policy based on this feedback to maximize the long-term reward.

### 2. Technical terms

Although among the three types of learning (supervised, unsupervised, and reinforcement learning), it is the most natural and closest to how humans learn, RL is particular in the sense that it has many unique jargons and vocabulary. In this sub-section, we aim to elucidate the main components of the RL setting.

As illustrated in Figure 1, the RL setting is composed of an **agent** (a learning-based model) that is interacting with an **environment** (e.g., simulation such as a game, systems such as an autonomous vehicle or drone, processes such as scheduling program, or even life itself, etc.). The environment receives the agent's **action** and changes its **state** accordingly, and this state is generally associated with a specific **reward**. For example, in an autonomous driving problem, when the car is in a safe and legal position, that is a state which the agent will receive a positive reward for. If the car is in an illegal position or in an accident, on the other hand, the agent will receive a negative reward, also called punishment. The sequence of actions executed by the agent given the corresponding environment's states is referred to as the **policy**.

All RL and deep RL algorithms must ensure an optimal balance between exploration and exploitation, which is a concept that is particularly associated with RL. It states that agents should explore the environment more in the first learning episodes, and progressively converge towards exploiting the learned knowledge that enables making optimal decisions. This is important because, if the agent starts by exploiting/using a learned policy from the beginning, it might miss other policies of higher optimality, thus, exploration is necessary. Mathematically, these components are all gathered in one framework known as the Markov decision process (MDP) [1], as {S, A, T, R, γ}. Where S is the state, A is the action, T is the transition function describing the dynamics of the environment, R is the reward function, and γ is a discount factor. The discount factor determines how much the RL agent cares about rewards in the immediate future compared to those in the distant future. This idea is borrowed from economics, where a certain amount of money is generally worth more now than in the distant future, also the discount factor helps in mathematical convergence. Solving this MDP refers to finding an optimal policy ($\pi^*$), that is the one yielding the maximum rewards over an entire learning episode.

Following the process depicted in Figure 1, first, the agent is presented with an initial state of the environment $s_t = s_0$, the reward associated with the initial state is usually considered null, i.e., $r_t=r_0=0$. The agent generates an action $a_t$ the state $s_t$. This action changes the environment's state to $s_{t+1}$, as well as the new associated reward $r_{t+1}$. The cumulated sum of rewards is known as the return $G_t$ and can be calculated as described in equation 1:

$$G_t = R_{t+1} + R_{t+2} + R_{t+3} + \ldots \quad \text{we begin by } R_{t+1} \text{ since } R_t \text{ is considered 0} \tag{1}$$

Then, we introduce the discount factor gamma γ

$$G_t = \gamma^{(0)} R_{t+1} + \gamma^{(1)} R_{t+2} + \gamma^{(2)} R_{t+3} + \ldots \quad \text{since } \gamma^{(0)} = 1, \text{ the equation becomes:}$$
$$G_t = R_{t+1} + \gamma^{(1)} R_{t+2} + \gamma^{(2)} R_{t+3} + \ldots = \sum_{k=0}^{T} \gamma^{(k)} R_{t+k+1} \tag{2}$$

Besides its role in modeling the importance of future rewards relative to immediate ones, the parameter γ gamma also ensures the mathematical convergence of the RL training. For instance, if γ = 0.99 (which is usually the value given in the literature 1<γ<0), then:
$G_t = R_{t+1} + 0.99^{(1)} R_{t+2} + 0.99^{(2)} R_{t+3} + \ldots = \sum_{k=0}^{T} 0.99^{(k)} R_{t+k+1}$, while the powers of γ tend to 0 (i.e., $\gamma^{(1)}$ =0.99, …, $\gamma^{(3)}$ =0.97, …, $\gamma^{(10)}$ =0.90, …, $\gamma^{(T)}$ ~0).

In addition to the discount factor added to the equation of the return $G_t$, to make this setting even more suitable for real case scenario, we shall consider a stochastic version of equation 2, that is, the expected return of $G_t$ also called the value function $V_t$, described in equation 3:

$$V_t = E[G_t | s_t = s] = E[\sum_{k=0}^{T} \gamma^{(k)} R_{t+k+1} | s_t = s] \tag{3}$$

Given the mathematical expectation in equation 4:
$$E[X] = \sum_{i=0}^{n} x_i P(x_i) \tag{4}$$

where $x$ is a random variable and $P$ is the probability of $x$.

Solving the MDP now consists of maximizing the value function (equation 3). Nevertheless, this is still a difficult problem, indeed, this is known as the infinite horizon problem since rewards are completely untraceable at a certain level. For this, we introduce the infamous **Bellman equation** trick. In the 1950s, Richard Bellman et al. [1] proposed a recursive approximation of the return $G$ and the value $V$ as follows:
We know that: $G_t = R_{t+1} + \gamma^{(1)} R_{t+2} + \gamma^{(2)} R_{t+3} + \ldots$
This can be rewritten as the immediate reward $R_{t+1}$ plus the discounted future values: $G_t = R_{t+1} + \gamma^{(1)} G_{t+1}$ and so on for later returns.
Similarly, the value function $V_t$ can also be arranged in this way:

$$V_t = R_{t+1} + V_{t+1} \tag{5}$$

This recursive representation helps avoid the infinite horizon problem by dividing it into solvable sub-problems. Later, this became known as the Bellman equation.

Based on the concepts and ideas elaborated above, many algorithms and techniques were proposed to solve such reinforcement learning problems. In Table 1 we present a subset of the foundational RL and deep RL algorithms with a taxonomy of characteristics.

*Table 1: Taxonomy of the RL and DRL algorithms*

| Algorithm | Value-based Policy-based | Model-based Model-free | On-policy Off-policy | Temporal-diff. Monte Carlo | Tabular NN |
|---|---|---|---|---|---|
| Q-learning | 0 | 1 | 1 | 0 | 0 |
| DQN | 0 | 1 | 1 | 0 | 1 |
| REINFORCE | 1 | 1 | 0 | 1 | 1 |
| DDPG | 0 and 1 | 1 | 0 and 1 | 0 | 1 |
| TD3 | 0 and 1 | 1 | 0 and 1 | 0 | 1 |
| PPO | 0 and 1 | 1 | 0 and 1 | 0 | 1 |

The following are definitions of the terms used in the taxonomy presented in Table 1:
- Value-based: updates the model's policy based on the value function $V$.
- Policy-based: updates the model's policy directly without referring to $V$.
- Model-based: has access to the environment's dynamics (e.g., rules of chess).
- Model-free: does not have access to the environment's dynamics.
- On-policy: uses a model for exploration, and the same for exploitation.
- Off-policy: uses a model for exploration, and a separate one for exploitation.
- Temporal difference: updates the model each time step.
- Monte Carlo: updates the model using a statistical description (e.g., the mean value) of many time steps (often, the entire episode)
- Tabular: uses a table to store the computed values
- Neural network (NN): uses a neural network as the function approximator

In the following section, we shall present in detail the algorithms listed in Table 1, each with the motivation behind its development, its inner working, as well as some of its known limitations. These algorithms are all model-free, leaving model-based algorithms for future work.

## 3. Algorithms: motivation, inner workings, limitation

### 1. Q-learning (value-based)

#### 1. Motivation

Before the NN and DL revolution, tremendous research was conducted to propose efficient techniques for solving the Bellman equation in the context of RL. For this purpose, there have been three principal classes of methods: dynamic programming, Monte Carlo methods, and temporal-difference (TD) learning. Although dynamic programming methods are well-developed mathematically, they must be provided a complete and accurate model of the environment, which is often not available. Monte Carlo methods on the other hand do not require a model and are conceptually simple but are not suited for step-by-step incremental computation (i.e., only episodic training). Finally, TD methods require no model and are fully stepwise incremental and episodic as well. Consequently, we will see that most algorithms are either TD-based (mostly) or Monte Carlo-based (less often), but rarely based on dynamic programming. Thus, to illustrate the RL paradigm here, we present the Q-learning algorithm, which is a TD-based model that belongs to the tabular RL model family. Other RL algorithms (i.e., do not rely on NNs), such as the State-action-reward-state-action (SARSA) which is an on-policy version of Q-learning, and the Monte Carlo Tree Search (MCTS) that relies on heuristic trees instead of a finite size table, can be found in Sutton & Barto's book [1].

## 2. Inner-workings

Q-learning (founded in 1989 by Christopher Watkins) is one of the basic, yet classic algorithms. Instead of the value function $V$ presented previously, Q-learning considers a variant of this, called $Q$-function, where Q designates quality. While $V$ is only associated with states, $Q$ is conditioned on states and actions as well.

$$Q_t = E[Q_t | s_t = s, a_t = a] = E[\sum_{k=0}^{T} \gamma^{(k)} R_{t+k+1} | s_t = s, a_t = a] \qquad (6)$$

This makes the evaluation of the agent's policy easier and explicit, and since we can utilize the mathematical expectation formula (equation 4), the relationship between $V$ and $Q$ is formulated as:

$$V^\pi = \sum_{a \epsilon A} \pi(a|s) Q^\pi(s, a)$$

As illustrated in Figure 2, Q-learning uses Q-table to store the Q-value of taking a certain action in a given

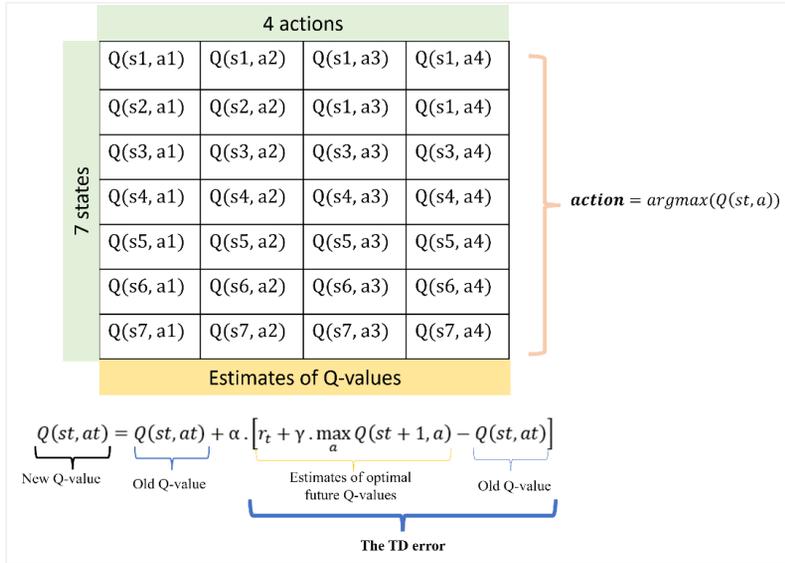

Figure 2: Q-learning

state. Then it updates the Q-value based on the observed reward and the estimated future value.

$$Q(st, at) = Q(st, at) + \alpha \cdot \left[ r_t + \gamma \cdot \max_a Q(st+1, a) - Q(st, at) \right] \qquad (7)$$

After running many iterations of the algorithm described in **Algorithm 1** below, the model (i.e., the Q-table) should converge, meaning that the highest Q-values will be assigned to the optimal state-action pairs. If interested, see the proof of convergence here [24].

Note for algorithm 1:
- s' and a' means next state and next action, respectively.
- ε-greedy is a method to ensure the exploration-exploitation process. Here, ε is set to a small number (<0.5), and a number $x$ is randomly generated, if $x < \varepsilon$, the algorithm explores, that is, it selects actions randomly. Whereas, if $x > \varepsilon$, the algorithm exploits, that is, it uses the Q-table to sample the best action, i.e., the one with the highest Q-value.

| **Algorithm 1:** Q-learning |
|---|
| Initialize **Q(s, a)** arbitrarily |
|     Repeat for each episode: |
|     Initialize **s** |
|     Repeat each step of the episode: |
|         Choose **a** given **s** using the policy derived from **Q** (e.g., ε-greedy) |
|         Take action **a**, observe **r, s'** |
|         Q(s, a) <= Q(s, a) + $\alpha \, [r + \gamma \, max \, Q(s', a') - Q(s, a)]$ |
|         **s** <= **s'** |
|     Until **s** is terminal |

### 3. Limitation

Although it was standard in the past years before the NNs revolution. Q-learning algorithm can only be applied in limited applications. This is because of its main drawback, which is its inability to (directly) learn in environments with continuous states and actions since it uses finite-size tables as brains.

## 2. Deep Q-learning (value-based)

### 1. Motivation

Given that in the previous sub-section, we laid out many concepts related to RL in general and the Q-learning algorithm specifically. understanding the deep version of the latter, that is, Deep-Q-network (DQN) should be relatively easy.

All concepts related to Q-learning (e.g., Bellman equation, ε-greedy exploration, etc.) still apply to DQN, except for one fundamental difference, which is the use of function approximators, specifically, NNs instead of tabular storage for the Q-values. In 2013, Mnih et al. from DeepMind introduced the DQN architecture. The work was labeled revolutionary as they used DQN to play a famous Atari game at the super-human level. Since then, DQN was used in many applications including medicine [25], industrial control and engineering [26], traffic management [27], resource allocation [28], as well as games [3,4]. Moreover, DQN work has opened the doors for many theoretical improvements and DRL algorithms developments, and to date, the DQN paper has been cited more than 9500 times.

### 2. Inner-workings

DQN has recognized two successive architectures, where the second one imported significant improvement to the first. The former DQN architecture involved a naïve straight-forward replacement of the Q-table in Q-learning by a NN, enabling a gradient-based update of the TD loss instead of the traditional dynamic programming approach. While this has worked for a few easier tasks, it suffered from two main problems, namely:

- The experience correlation: when the agent learns from consecutive experiences as they occur, data can be highly correlated. This is because a given state may provoke a certain action which in turn will provoke a specific state and action and so on, given those environments are governed by well-defined transition functions.
- The moving target problem: unlike supervised learning (SL), where data is fixed, in RL, the agent generates its data, and the more the agent changes its policy, the more the data distributions change, which makes calculating the Q-values very unstable. Therefore, the model is said to be chasing a moving target. This is because each time the Q-network is updated, and so is the Q-value will be changing constantly. Formally, if the update is computed for $rt + \max_{a} Q(st + 1, a; \theta) - Q(st, at; \theta)$, then the learning is prone to becoming unstable since the target,

$rt + \max_a Q(st+1, a; \theta)$, and the prediction $Q(st, at; \theta)$, are not independent, as they both rely on θ.

For this, the second architecture was proposed and demonstrated significant mitigation of the two problems discussed. This was achieved by adding two main blocks to the original architecture:
- The replay memory: the replay memory, experience replay, or even the replay buffer. This helps break the correlation between successive experiences and ensures *independent, identical, distributed* (IID)-like data which is assumed in most supervised convergence proofs. It achieves this by storing transitions of {state, action, reward, new state} until having batches of these. Then, randomly sampling transitions for learning in a phase that is separate from the interaction with the environment/experience generation phase.
- The target network: by setting a separate network for the calculation of the target Q-value, the model becomes more robust and stable. It is because only after batches of learning episodes (as opposed to each timestep in the former case) that the target can change due to the update. Thus, the model can succeed in approximating the target progressively.

Illustrated in Figure 3 is the enhanced version of the DQN architecture. Along with **algorithm 2**, we shall explain its inner working. After initializing the two networks' parameters (Q-network with *θ* and target network with *θ'*, where *θ= θ'*), the learning episodes begin. We first initialize the initial state and begin each episode step by step following the RL general workflow. The agent gives an action based on the state provided, and the environment receives the action and outputs a new state along with the corresponding reward. This (so-called) transition {state, action, reward, new state} is stored in a memory (i.e., the replay memory). After a couple of episodes, the stored transitions are sampled randomly from the replay memory (to avoid correlation of experiences) to compute their corresponding Q-values. The Q-value given by the Q network is the predicted one, whereas the Q-value given by the target network with the Bellman approximation is considered the real one. Finally, the Q network is updated via the mean-square error (MSE) to minimize the loss between the predicted Q and the real one. Whereas the target network's parameters are updated by copying the Q network's parameters, which helps avoid the moving target problem.

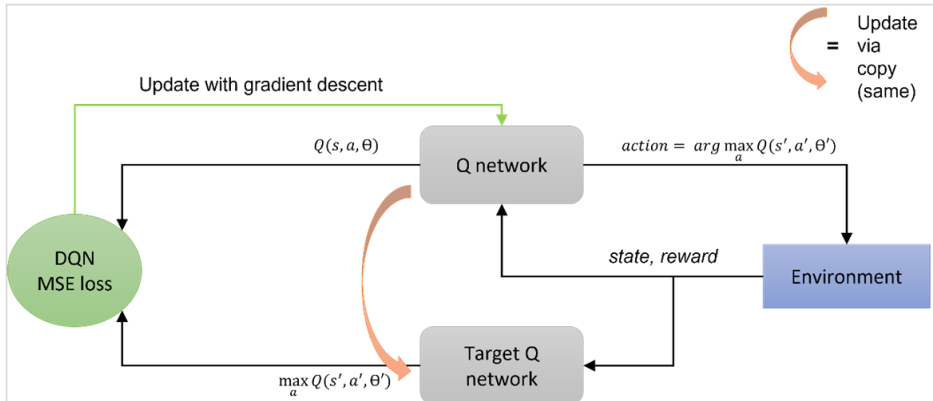

*Figure 3: Deep Q-Network (DQN)*

**Algorithm 2:** DQN

Initialize replay memory $D$ to capacity $N$
Initialize action-value function $Q$ with random weights $\theta$   #***NN***
Initialize target action-value function $\widehat{Q}$ with weights $\theta' = \theta$  #***NN***
**For** episode = 1, $M$ **do**
    Initialize sequence $s_1$
    **For** t = 1, T **do**
        With probability, $\varepsilon$ select $a$ random action $a_t$
        Otherwise select $\boldsymbol{a_t} = \boldsymbol{argmax\ Q(s_t,a;\theta)}$
        Execute action $\boldsymbol{a_t}$ and observe reward $r_t$ and next state $s_{t+1}$
        Store transition $\boldsymbol{s_t, a_t, r_t, s_{t+1}}$ **in $D$**
        Sample a random minibatch of transitions from D
        Set $\boldsymbol{y_j} = \begin{cases} r_j, & \textit{if episode terminates at step } j+1 \\ r_j + \gamma max \widehat{Q}(s_t + 1, a'; \theta'), & \textit{otherwise} \end{cases}$
        Gradient descent on $\boldsymbol{MSE\ (y_j, Q(s_t, a; \theta))}$, with respect to $\theta$
        For every C step, reset $\widehat{Q} = Q$
    **End for**
**End for**

### *3. Limitation*

As mentioned, DQN has been driving a tremendous number of applications in many domains. Nevertheless, it is still limited, mainly, because of the categorical way by which actions are generated ($a_t = argmax\ Q(s_t, a; \theta)$). This makes it unable to be used in continuous action settings (e.g., controlling velocity). On top of this, DQN has also shown significant training instability in high-dimensional environments where function approximation of the Q-value can become prone to significant overestimation error [29].

## 3. REINFORCE (policy-gradient)

### *1. Motivation*

Policy gradient (PG) algorithms are the most natural and direct way of doing RL with neural networks. This family of RL and DRL techniques was popularized by Sutton et al. in [20]. It proposes a new set of algorithms that can directly map states to actions (as opposed to Q-values, then actions) via a learnable differentiable function approximator such as NNs. It should be differentiable since the update is based on the gradient.

### *2. Inner-workings*

Unlike value-based methods such as the previously presented Q-learning and DQN, PG algorithms have a much more solid mathematical basis. Indeed, instead of the Bellman approximation of the Q-values via the recursive decomposition. The gradient loss in PG algorithms is directly proportional to the policy in an intuitive way. Below is the PG theorem proof:

We know that the mathematical expectation $E$ is equal to:

$E[X] = \sum_{i=0}^{n} x_i P(x_i)$, where $x$ is the random variable and $P$ is the probability of $x$. therefore, the gradient of the objective function: $\nabla_\theta J(\pi_\theta) = \nabla_\theta \mathbb{E}_{\tau \sim \pi_\theta} [R(\tau)]$ (where $\tau$ is the trajectory) can be formulated as:

$$= \nabla_\theta \int_\tau P(\tau|\theta) R(\tau), \quad \quad \text{expanding the expectation formula}$$

$$= \int_\tau \nabla_\theta P(\tau|\theta)R(\tau), \quad \text{multiply and divide by highlighted term}$$

$$= \int_\tau \frac{\nabla_\theta P(\tau|\theta)R(\tau)P(\tau|\theta)}{P(\tau|\theta)}, \quad \text{the highlighted part is } \frac{du}{u} = d\log(u)$$

$$= \int_\tau P(\tau|\theta)\nabla_\theta \log(P(\tau|\theta))R(\tau), \quad \text{log derivative trick}$$

$$= E_{\tau \sim \pi_\theta}[\nabla_\theta \log(P(\tau|\theta))R(\tau)], \quad \text{original expectation form}$$

Among the terms in the last expression, only $P(\tau|\theta)$ needs to be defined (reward is known). Since $P(\tau|\theta)$ the probability of following a trajectory $\tau$ is dictated by the policy function $\pi_\theta$, thus, the final expression becomes:

$$\nabla_\theta J(\pi_\theta) = E_{\tau \sim \pi_\theta}\left[\sum_{t=0}^{T} \nabla_\theta \log(\pi_\theta(a_t|s_t))R(\tau)\right], \tag{8}$$

By performing the gradient ascent on this by a Monte Carlo estimate of the expected value, we can find the optimal policy. While many tricks were proposed to make the PG algorithm better, all improved PG algorithms are still based on this basic theorem.

This final expression allows the easy and direct gradient-based update of the agent's parameters solely based on its policy (i.e., actions) instead of the indirect way, i.e., state-action value (Q-value)). Moreover, it permits the training of RL agents in continuous setting environments naturally. This is because the action selection is not based on a categorical sampling method such as the *argmax* in DQN. Most PG-based algorithms can be used for discrete action environments, via any categorical sampling function, as well as the continuous settings, where the agent tries to learn a distribution function such as the gaussian distribution conditioned on its parameters, the mean (μ), and the variance σ.

The simplest PG algorithm is called REINFORCE. Here, the model follows a naïve strategy of using a NN to map states to actions optimally (i.e., in a way that maximizes the objective function). In other words, The REINFORCE algorithm presents a direct differentiation of the reinforcement learning objective, i.e., the expected return described in equation 3: $E[\sum_{k=0}^{T}\gamma^{(k)}R_{t+k+1}|s_t = s]$. As illustrated in Figure 4 and presented in **algorithm 3** there is only one network, which is the policy network ($\pi$). The agent learns to choose optimal actions given the environment's states to increase the objective function (in equation 8) via gradient ascent.

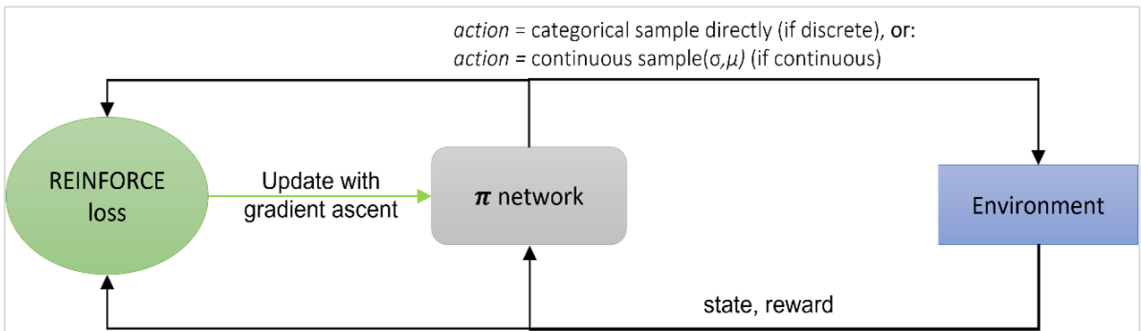

*Figure 4: REINFORCE*

**Algorithm 3:** REINFORCE (policy function is a NN)

---

Initialize a differentiable policy function $\pi(a|s, \theta)$  #*NN*
Repeat for each episode:
    Record $S_0, A_0, R_1, ..., S_{T-1}, A_{T-1}, R_T$, following $\pi$
    Loop for each step of the episode $t = 0, 1, ..., T-1$
        $G_t \leftarrow \sum_{k=t+1}^{T} \gamma^{k-t-1} R_k$
        $\theta \leftarrow \theta + \alpha \gamma^t G_t \nabla ln\pi(A_t|S_t, \theta)$  #$\pi(A_t|S_t, \theta)$ will return the probability of $A_t$

---

### 3. Limitation

With all the advantages provided by the PG theorem and its direct implementation. The REINFORCE algorithm suffers from fundamental problems, namely, the noisy estimates of the gradient caused by the unclear credit assignment. That is, it is not clear which action resulted in which reward. This is because, as described in algorithm 3, the agent is only updated when a complete episode of data (i.e., all steps) or more is completed. This is necessary since PG is based on a statistical Monte Carlo-like estimate of the cumulated reward samples (return) generated by the policy.

For this, a critical improvement was proposed as a new family of DRL algorithms known as actor-critic architectures [30]. To date, the actor-critic-based models are the SOTA in the DRL field, especially, in the model-free DRL.

The key contribution of actor-critic models is the use of both the policy and the value for learning optimal strategies. This is the reason why in Table 1, all models after REINFORCE were dubbed value-based as well as policy-based. In the actor-critic architecture, the actor is a NN that is responsible for learning the policy by trying to maximize an objective function similar to that of the REINFORCE algorithm. Whereas the critic (also a NN) learns the value function $V$, the action value function $Q$, or even a relationship between the two, called the advantage $A$, which is just the difference between the next estimated value (V or Q) and the previous one as described in equation 17 below:

$$A_t = \begin{cases} r_t + \gamma V(s_{t+1}) - V(s_t), & or: \\ r_t + \gamma Q(s_{t+1}, a_{t+1}) - Q(s_t, a_t) \end{cases} \quad (9)$$

In the following subsections, we will overview three SOTA models that were making headlines in the past five years or so. These are all actor-critic-based models, so more details about this family of DRL algorithms will be presented in each sub-section.

### 4. DDPG (actor-critic):

#### 1. Motivation

Deep deterministic policy gradient, or in short DDPG, was developed in 2016 by Lillicrap et al. to have a DRL model that can operate in a high dimensional continuous state and action space. DDPG, as the name suggests, uses a deterministic policy, as opposed to a stochastic one, by mapping states to actions directly instead of a probability distribution over actions via a deterministic policy function named $\mu(s|\theta^\mu)$.

#### 2. Inner-workings

As shown in Figure 5, DDPG is an actor-critic model, meaning that it contains at least two networks, one called the actor responsible for learning the policy, and a second network named the critic for learning the value function of each state ($V$) or state-action pair ($Q$). The critic learns by the mean of minimizing the MSE loss between the predicted Q-value and the actual one akin to the DQN model discussed previously. As the critic gets better at predicting the corresponding Q-values, the actor is presented with a convenient value that it can use to maximize in a PG fashion. Further, DDPG makes use of other improvements made for the DQN model, namely, the replay memory and the target network.

In **algorithm 4** we quote the line "Store transition $(s_t, a_t, r_t, s_{t+1})$ in $D$" which refers to the same replay memory used in the DQN model. This is because it was shown that PG models, including those with actor-critic architectures, are very sample inefficient since they only use a given transition once and then throw it off. The replay memory allows training in mini-batches of data as well as using the same data multiple times during training. Another change is the way the target is updated, while DQN updates the target network by the mean of a hard copy of the parameters, DDPG uses a soft copy based on the Polyak averaging function as follows $\theta^{Q'} \leftarrow \tau\theta^Q + (1-\tau)\theta^{Q'}$. As we can see, the new parameters of the target will equal an interpolation factor $\tau$ which is usually set to a value around 0.99, plus a small number (~1 −0.99, i.e., 0.01) times the old parameters of the target network. This constrains the target network to be changed slowly and was shown (in the DDPG paper) to yield better performance than the hard copy of the DQN model.

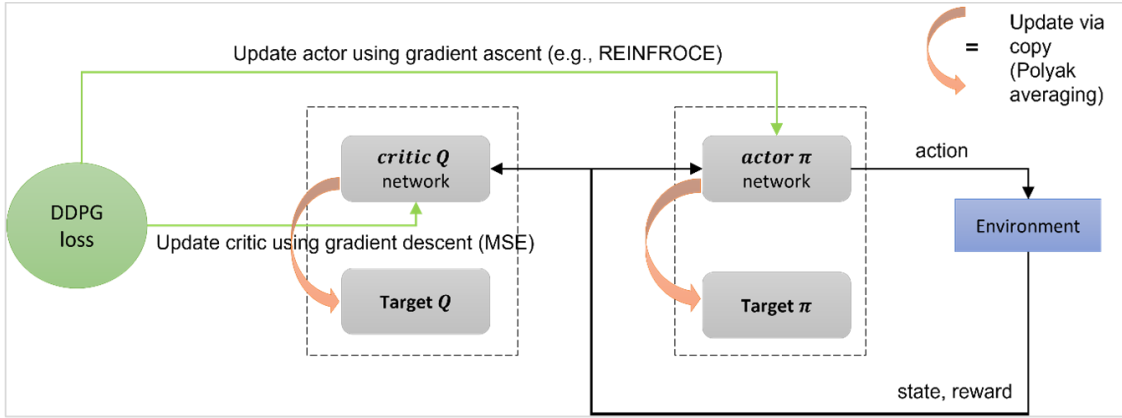

*Figure 5: Deep deterministic policy gradient (DDPG)*

Further, since the DDPG model is merely deterministic, it must ensure the exploration-exploitation somehow. Indeed, the action selection is executed according to the equation mentioned in the algorithm, specifically: $a_t = \mu(s_t|\theta_\mu) + N_t$, where $N_t$ is a noise signal added to the policy function to force the stochastic behavior. In the paper, the researchers used the Ornstein Uhlenbeck noise [31], however, later research has demonstrated that other simple noise signals such as the gaussian noise can be used and yield competitive performance.

**Algorithm 4:** DDPG

Initialize the critic and the actor networks, $Q(s, a|\theta^Q)$ and $\mu(s|\theta^\mu)$
Initialize target network $Q'$ and $\mu'$, $\theta^{Q'} \leftarrow \theta^Q, \theta^{\mu'} \leftarrow \theta^\mu$
Initialize replay memory R
**For** episode = 1, M **do**
    Initialize a random process N for action exploration
    Receive initial observation state s1
    **For** t = 1, T do
        Select action $a_t = \mu(s_t|\theta_\mu) + N_t$ #policy + noise
        Execute at, observe reward rt and new state st+1
        Store transition $(s_t, a_t, r_t, s_{t+1})$ in $D$
        Sample a random minibatch of K transitions from $D$
        Set $yi = r_j + \gamma Q'(s_{i+1}, \mu'(s_{i+1}|\theta^{\mu'}) | \theta^{Q'})$
        Update critic by minimizing the loss: $MSE(y_i, Q(s_i, a_i|\theta^Q))$
        Update the actor by maximizing the expected $Q$:
$$\nabla_{\theta^\mu} J \approx \frac{1}{K} \sum_{i=0}^{K} \nabla_a Q(s, a|\theta^Q)|_{s=s_i, a=\mu(s_i)} \nabla_{\theta^\mu} \mu(s|\theta^\mu)|_{s_i}$$
        Update the target networks:
            $\theta^{Q'} \leftarrow \tau\theta^Q + (1-\tau)\theta^{Q'}$    #Polyak averaging
            $\theta^{Q'} \leftarrow \tau\theta^\mu + (1-\tau)\theta^{\mu'}$
    **End for**
**End for**

### 3. Limitation

While it has empowered several great achievements in the AI field, DDPG is often criticized for being unstable. This is manifested in the form of high sensitivity to hyperparameter tuning and propensity to converge to very poor solutions or even diverge in tasks requiring relatively more exploration [32]. For this, potential improvements were open to further the DRL decision-making optimality, especially for high-dimensional spaces and noisy environments.

## 5. TD3 (actor-critic):

### 1. Motivation

The limited performance of the DDPG model was extensively explored. As a result, the cause was identified to be *the overestimation bias*. This problem was already known as an issue regarding Q-learning. It is manifested when the estimated values Q-values are often greater than true ones, thus, the agent is said to be overestimating the expected future rewards, making it select actions of misleading high values. This phenomenon accumulates after each iteration and propagates through the TD error.

The overestimation bias is generally related to the statistical-based approximation via NNs or another approximative sample-based method such as dynamic programming decomposition in Q-learning. As a natural successor to DDPG, and given its main limitations discussed. The twin-delayed deep deterministic (TD3) PG model was developed in 2018.

### 2. Inner-workings

TD3 solves the overestimation bias in the DDPG model issue by maintaining a pair of critics Q1 and Q2 as illustrated in Figure 5 (hence the name "twin") along with a single actor, and each network with its corresponding target.

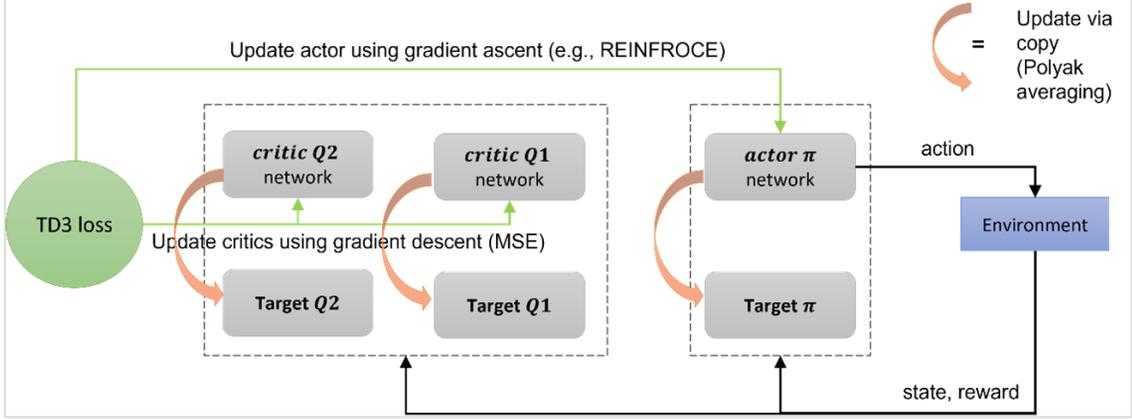

*Figure 6: Twin delayed deep deterministic (TD3) policy gradient*

For each time step, TD3 uses the smaller of the two Q-values and updates the policy less frequently than the critic networks. It is worth mentioning that the use of two networks for better approximation of the Q-value was inspired by previous work concerning the overestimation bias in Q-learning and DQN, namely, double Q-learning and double DQN [29], respectively. Besides these changes, the TD3 model is akin to the DDPG model in both selecting actions and updating the networks as explained in **algorithm 5**.

| **Algorithm 5:** TD3 |
|---|
| Initialize the critic networks $Q_{\theta 1}$, $Q_{\theta 2}$, and the actor-network $\pi_\Phi$, randomly |
| Initialize the target networks $\theta_1` \leftarrow \theta_1$, $\theta_2` \leftarrow \theta_2$, $\Phi` \leftarrow \Phi$ |
| Initialize the replay memory $B$ |
| **For** t = 1 to T **do** |
|     Select action with exploration noise $a \sim \pi(s) + \varepsilon$, |
|     $\varepsilon \sim N(0, \sigma)$ and observe reward r and new state **s'** |
|     Store transition *(s, a, r, s')* in $B$ |
|     Sample a mini-batch of N transitions from $B$ |
|     $\tilde{a} \leftarrow \pi_\Phi'(s) + \varepsilon$, #**policy + noise** |
|     $y \leftarrow r + \gamma \min_{i=1,2} Q_\theta'(s', \tilde{a})$ #**Clipped double Q-learning** |
|     Update critics $\theta_i \leftarrow \min_{\theta_i} MSE(y, Q_{\theta_i}(s,a))$ |
|     **If** t mode *d* then  #Delayed update of target and policy networks |
|         Update $\Phi$ by the deterministic policy gradient: |
|         $\nabla_\Phi J(\Phi) = N^{-1} \sum_{i=0}^{K} \nabla_a Q_{\theta_1}(s,a)\|_{a=\pi_\Phi(s)} \nabla_\Phi \pi_\Phi(s)$ |
|         Update target networks: |
|             $\theta'_i \leftarrow \tau\theta Q + (1 - \tau)\theta'_i$  #Polyak averaging |
|             $\Phi' \leftarrow \tau\Phi + (1 - \tau)\Phi'$ |
|     **End if** |
| **End for** |

3. Limitation

The exploration is handled manually with a gaussian noise, and the use of six networks is computationally heavy. Further, TD3 can only be used naturally (if no features engineering is considered) with continuous space environments.

6. PPO (actor-critic):

1. Motivation

In contrast to DDPG and TD3, where the exploration is executed by manually introducing a noise parameter to the policy. There exists another category of actor-critic PG algorithms that uses a stochastic policy, as opposed to the deterministic one. This inherently makes the exploration phase handle automatically. Among the main SOTA models that utilize this approach is the proximal policy optimization (PPO) algorithm.

2. Inner-workings

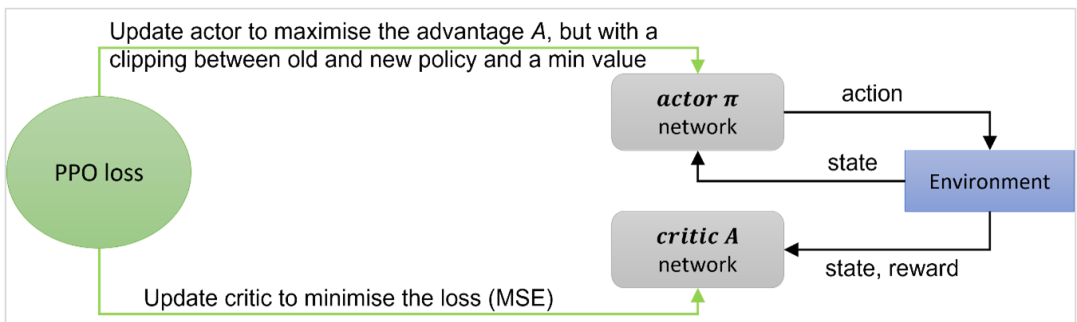

*Figure 7: Proximal policy optimization (PPO)*

Similar to the REINFORCE model, PPO learns online, meaning that it collects mini batches of experience while interacting with the environment and use them to update its policy. Whereas DDPG and TD3 are considered off-policy (or off-line) models since the function used to collect experiences (*argmax*) is different from the one updated for learning ($\pi$).

Once the PPO policy is updated, a newer batch is collected with the newly updated policy. The critical issue introduced by PPO is the concept of *trust region*, which refers to the fact that a new update of the policy should not change it too much from the previous policy. This resulted in significantly less variance and smoother training and makes sure the agent does not go down an unrecoverable path. Indeed, we quote the original paper: "These methods have the stability and reliability of trust region policy optimization (TRPO) [33] methods but are much simpler to implement, requiring only a few lines of code change to a vanilla policy gradient implementation"

The PPO model was developed for three main points:
- ease of implementation
- sample efficiency
- ease of tuning

As one can tell from Figure 5, although the TD3 model has shown impressive improvement compared to previous models, it involves many complexities regarding the implementation and training, especially, its use of six separate networks, as opposed to the architecture of PPO illustrated in Figure 6.

Additionally, the main problem of the previously presented on-policy model (REINFORCE) suffered from the sample inefficiency since it only uses the collected transition once. Here, PPO is an on-policy model, yet it can also utilize the same transition multiple times by using a clipping function, known as the surrogate loss

(mentioned in **algorithm 6**), to force the ratio of the old policy and the new one to stay between a manageable interval. This key contribution of the PPO model is also helpful for hyperparameters tuning, which can be very laborious for DDPG and TD3 given the complex architectures with several parts as well as the risk of going outside the trust region, especially, if the learning rate is larger than it should be.

Formally, PPO has imported small, yet very effective changes to the simple PG form such as the REINFORCE model. That is, by replacing the "log $(\pi_\theta(a_t|s_t))$" in the PG objective function of equation 16, recall:

$$\nabla_\theta J(\pi_\theta) = E_{\tau \sim \pi\theta}\left[\sum_{t=0}^{T} \nabla_\theta \log(\pi_\theta(a_t|s_t))R(\tau)\right]$$

by the ratio $r_t(\theta) = \frac{\pi_{\theta new}}{\pi_{\theta old}}$, clipping it between $1 - \epsilon$ and $1 + \epsilon$, and maximizing it using gradient ascent just like a normal PG algorithm.

---

**Algorithm 6:** PPO

Initialize policy parameters $\theta_0$, initialize value function parameters $\Phi_0$
**For** k= 0, 1, 2, ..., **do**
    Collect a set of trajectories $D_k = \{\tau_i\}$ by running policy $\pi(\theta_k)$ in the environment
    Compute rewards-to-go $\hat{R}_t$.
    Compute the advantage estimates, $\hat{A}_t = \delta_t + (\gamma\lambda)\delta_t + 1 + ... + (\gamma\lambda)T - t + 1\delta_T + 1$,
        where $\delta_t = r_t + \gamma V\Phi(s_t + 1) - V\Phi(s_t)$ #like TD in the Bellman equation
    Update the policy by maximizing the PPO-clip objective:
      **#Below is the clipped surrogate loss, where $r_t(\theta) = \frac{\pi_{\theta new}}{\pi_{\theta old}}$**
    $L_{\theta_k}(\theta) = E\left[\sum_{t=0}^{T}[\min(r_t(\theta)\hat{A}_t^{\pi_k}, clip(r_t(\theta), 1 - \epsilon, 1 + \epsilon)\hat{A}_t^{\pi_k}]\right]$
    Fit the value function: $V_\Phi(s_t, \hat{R}_t)$ via MSE loss
**End for**

---

*3. Limitation*

To date, PPO is still considered a SOTA algorithm, and its limitations are yet to be explored. It was created by a team at OpenAI, and it is still the main algorithm used in the company. Indeed, it was recently used to fine-tune their ChatGPT model as well.

## 4. Summary:

In this review, we have provided a comprehensive overview of the field of reinforcement learning (RL), covering the key concepts, techniques, and algorithms. We have traced the historical progress of the field, specifically, the main model-free algorithms, starting with the early Q-learning algorithm and covering the current state-of-the-art algorithms of deep RL such as DQN, TD3, and PPO (all discussed algorithms are listed in Figure 7 below).

We have also discussed the challenges and limitations of RL, including sample efficiency and the exploration-exploitation trade-off, and how these challenges have been addressed in the development of newer algorithms. One of the main advantages of RL algorithms is their ability to learn from experience and adapt to changing environments. This allows RL agents to solve complex and dynamic problems that may be difficult to model using other techniques. However, RL algorithms can also be computationally intensive and may require a large amount of data and interactions with the environment to learn effectively.

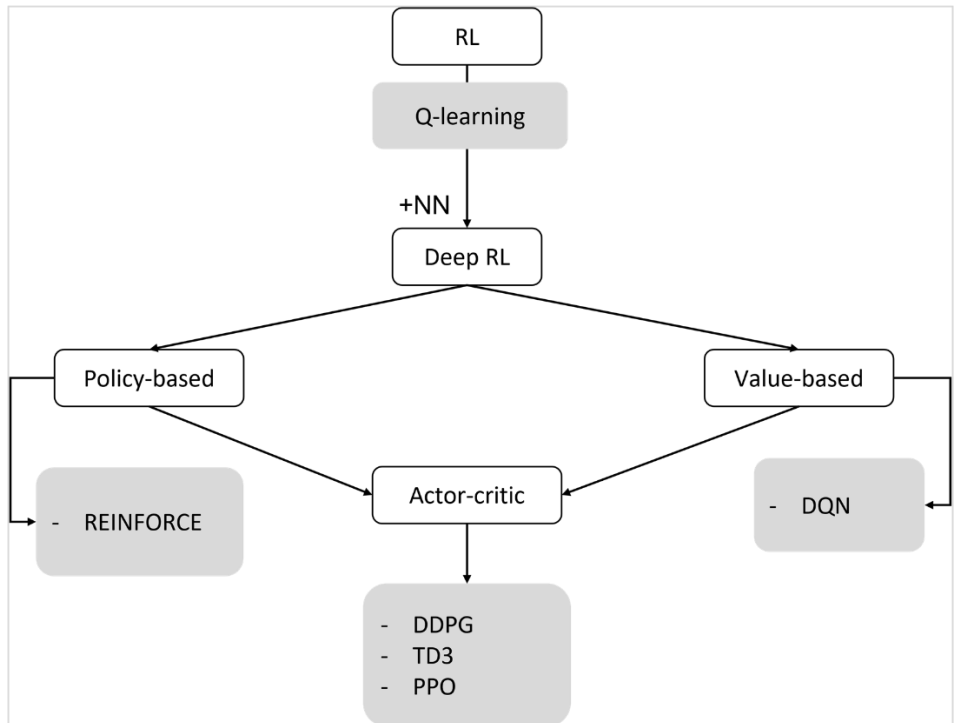

*Figure 8: Main RL-Algorithms discussed in the paper*

Given the diverse range of RL algorithms that have been developed, it can be challenging to choose the best algorithm for a particular problem. In general, model-based algorithms are more sample efficient but may be less robust than value-based or policy-based algorithms. Value-based algorithms are widely used and can be effective in many situations, but they may suffer from instability or divergence in certain cases. Policy-based algorithms are generally more stable and can handle high-dimensional action spaces, but they may be slower to converge compared to value-based algorithms.

In conclusion, the choice of the RL algorithm will depend on the specific characteristics of the problem at hand and the trade-offs between sample efficiency, stability, and convergence speed. This review paper aims to provide a comprehensive and accessible overview of the key concepts, techniques, and algorithms of RL, as well as the challenges and limitations of the field and its historical progress. It is intended to serve as a valuable resource for researchers and practitioners interested in learning more about RL and its applications.

## Acknowledgement:
None.None.

Algorithms, (2017) 1–12. http://arxiv.org/abs/1707.06347.